\documentclass[a4paper,fleqn]{cas-dc}

\usepackage[numbers]{natbib}
\usepackage{booktabs}
\usepackage{longtable}
\usepackage{multirow}
\usepackage{graphicx}
\usepackage{colortbl} 

\newcommand{\monaifunc}[1]{\texttt{#1}}
\newcommand{\monaiarg}[1]{\texttt{#1}}
\def\tsc#1{\csdef{#1}{\textsc{\lowercase{#1}}\xspace}}
\tsc{WGM}
\tsc{QE}
\tsc{EP}
\tsc{PMS}
\tsc{BEC}
\tsc{DE}

\begin{document}
\let\WriteBookmarks\relax
\def\floatpagepagefraction{1}
\def\textpagefraction{.001}
\shorttitle{Efficient Parameter Adaptation for Multi-Modal Medical Image Segmentation and Prognosis}
\shortauthors{N. Saeed et~al.}

\title [mode = title]{Efficient Parameter Adaptation for Multi-Modal Medical Image Segmentation and Prognosis}                      







\author[1]{Numan Saeed}[orcid=0000-0002-6326-6434]
\cormark[1]
\ead{Numan.Saeed@mbzuai.ac.ae}

\credit{Conceptualization of this study, Methodology, Software}

\affiliation[1]{organization={Department of Computer Vision, Mohamed bin Zayed University of Artificial Intelligence},
                city={Abu Dhabi},
                country={United Arab Emirates}}

\author[2]{Shahad Hardan}[]
\ead{Shahad.Hardan@mbzuai.ac.ae}

\author[1]{Muhammad Ridzuan}[]
\ead{Muhammad.Ridzuan@mbzuai.ac.ae}

\credit{Data curation, Writing - Original draft preparation}

\affiliation[2]{organization={Department of Machine Learning, Mohamed bin Zayed University of Artificial Intelligence},
                city={Abu Dhabi},
                country={United Arab Emirates}}

\author[1, 4]{Nada Saadi}[]
\ead{Nada.Saadi@mbzuai.ac.ae}

\author[1,3]{Karthik Nandakumar}
\ead{Karthik.Nandakumar@mbzuai.ac.ae}

\affiliation[3]{organization={Michigan State University},
                city={Michigan},
                country={United States}}

\affiliation[4]{organization={M42 Health},
                city={Abu Dhabi},
                country={United Arab Emirates}}

\author[1]{Mohammad Yaqub}
\cormark[2]
\ead{mohammad.yaqub@mbzuai.ac.ae}

\cortext[cor1]{Corresponding author}
\cortext[cor2]{Principal corresponding author}


\begin{abstract}
Cancer detection and prognosis relies heavily on medical imaging, particularly CT and PET scans. Deep Neural Networks (DNNs) have shown promise in tumor segmentation by fusing information from these modalities. However, a critical bottleneck exists: the dependency on CT-PET data concurrently for training and inference, posing a challenge due to the limited availability of PET scans. Hence, there is a clear need for a flexible and efficient framework that can be trained with the widely available CT scans and can be still adapted for PET scans when they become available. In this work, we propose a parameter-efficient multi-modal adaptation (PEMMA) framework for lightweight upgrading of a transformer-based segmentation model trained only on CT scans such that it can be efficiently adapted for use with PET scans when they become available. This framework is further extended to perform prognosis task maintaining the same efficient cross-modal fine-tuning approach. The proposed approach is tested with two well-known segementation backbones, namely UNETR and Swin UNETR. Our approach offers two main advantages. Firstly, we leverage the inherent modularity of the transformer architecture and perform low-rank adaptation (LoRA) as well as decomposed low-rank adaptation (DoRA) of the attention weights to achieve parameter-efficient adaptation. Secondly, by minimizing cross-modal entanglement, PEMMA allows updates using only one modality without causing catastrophic forgetting in the other. Our method achieves comparable performance to early fusion, but with only 8\% of the trainable parameters, and demonstrates a significant +28\% Dice score improvement on PET scans when trained with a single modality. Furthermore, in prognosis, our method improves the concordance index by +10\% when adapting a CT-pretrained model to include PET scans, and by +23\% when adapting for both PET and EHR data.



\end{abstract}




\maketitle
\section{Introduction}
Medical imaging plays a crucial role in modern healthcare by providing visual representations of the interior of the body. This is instrumental for the diagnosis, treatment, and monitoring of various medical conditions, such as cancer, neurological disorders, and cardiovascular diseases, among others \cite{Hussain2022Modern}. To achieve a comprehensive understanding, clinicians utilize a range of advanced imaging modalities, including CT, MRI, PET, and ultrasound, each offering unique and complementary perspectives on tissues and organs \cite{Dhermain2010Advanced, Hussain2022Modern, Langen2017Advances}. Despite this increasing utility and diagnostic power, medical imaging faces growing challenges, primarily related to increasing demand and interpretive workload. This is evidenced by historical trends: from 1999 to 2010, the number of cross-sectional imaging studies doubled, while the number of images increased tenfold. Furthermore, current estimates indicate over 80 million annual CT scans in the US \cite{harvard_radiation_risk_imaging}, with demand projected to surge further in the coming years, driven significantly by a projected 47\% increase in cancer incidence by 2040 \cite{Masjedi2019European, Sung2021Global}. Beyond the general volume increase, cancer patients frequently require multiple imaging examinations and ongoing monitoring throughout their treatment, adding significant strain to clinical resources \cite{Rehani2019Patients}. Automated lesion segmentation and tracking have the potential to partially alleviate this workload, supporting increasingly accurate and efficient assessments of tumor burden, disease progression, and radiomics analysis \cite{Rogers2020Radiomics:}. Although recent deep learning models have demonstrated strong performance in segmenting anatomical structures like organs \cite{bassi2024touchstone} using large labeled datasets \cite{ ji2022amos, 2020AbdomenCT-1K:}, lesion segmentation remains significantly more challenging \cite{du2024segvol, saeed2021ensemble, saeed2022tmss}. 

The development of hybrid imaging technologies has dramatically enhanced diagnostic capabilities by combining complementary modalities. PET/CT merges the metabolic information from PET with the anatomical detail of CT, enabling precise localization of functional abnormalities within anatomical structures. For instance, in head and neck cancer radiotherapy, 18F-FDG PET/CT-based gross tumor volume (GTV) delineation helps distinguish healthy from pathologic metabolic activity, though this task remains challenging when assessing nodal involvement or differentiating healthy metabolic activity in proximity to malignant tissue. It is well-known that integration of both CT and PET imaging modalities using DNNs can significantly enhance the accuracy of tumor segmentation \cite{Ren_2021}. Joint processing of these two imaging modalities has been extensively studied in the literature \cite{huang2024vision,saeed2021ensemble}. These approaches involve either merging the modalities at the input level as channels \cite{farag2021early} or at the output level after independent processing of each modality \cite{kadoury2011accuracy}. While the former approach assumes constant availability of both modalities, the latter approach doubles the number of trained parameters \cite{wang2022deep}. Hence, there is a need for more sophisticated method for jointly handling CT and PET scans.

Several challenges must be addressed when designing such a method. The first challenge involves the problem of datasets scarcity. It is well-known that performance of DNN-based medical image segmentation models is heavily impacted by the availability of large, multi-modal and specialized datasets \cite{li2025well}. However, the availability of PET/CT scans, combined, may be restricted in practice due to cost and clinical-expertise necessity. As a result, publicly available datasets are predominantly based on CT scans \cite{li2025well}. These include abdominal multi-organ datasets like BTCV (Beyond The Cranial Vault) \cite{igelsias2015miccai}, AMOS \cite{ji2022amos}, and AbdomenAtlas, to organ/tumor datasets, e.g., LiTS (Liver Tumor Segmentation) \cite{bilic2023liver}, KiTS (Kidney Tumor Segmentation) \cite{heller2019kits19}, and MSD (Medical Segmentation Decathlon) \cite{simpson2019large}. Hence, the proposed method should be able to efficiently adapt a pre-trained model on a commonly available modality to support additional or combined modalities.

The second challenge is that the method should allow prediction/segmentation when single imaging modality is available instead of multi-modal. For example, while PET/CT scans provide combined functional and anatomical insights, not all clinical settings have access to both modalities due to either cost or patient-specific needs. Thus, the method must perform robustly even when only CT or PET data is available.

The third challenge is that the method should continually adapt to the data of new medical centers. This requires the model to learn efficiently from new data over time without having access and losing performance on previously learned modalities and tasks. Existing works in continual learning often adopt an approach where the encoder and decoder are frozen, and an additional decoder is introduced to accommodate new classes \cite{ji2023continual}. This strategy, however, incurs significant memory costs due to the increased number of network parameters. Furthermore, the newly added decoder does not effectively leverage the relationships between the new organs or tumors and the existing classes, which restricts the use of prior knowledge. In our scenario, the challenge is amplified: we aim to incorporate not only new classes but also new modalities. This added complexity renders the simple addition of a new decoder insufficient. To address this, our objective is to design a network architecture that minimizes parameter growth across continual learning steps while ensuring the network remains aware of the relationships between the newly introduced modalities and the existing tasks.

To address these challenges, we develop a flexible framework, Parameter-Efficient Multi-Modal Adaptation for Medical Image Segmentation (PEMMA), that leverages efficient parameter fine-tuning for multi-modal adaptation as well as continual learning scenarios. Our approach is based on creating a base model that is trained on widely available CT scans and utilizes low-rank adaptation (LoRA) or Weight-Decomposed Low-Rank Adaptation (DoRA) of attention weights to efficiently incorporate PET scans or electronic health records (EHR) when they become accessible. PEMMA also supports continual learning, allowing the model to adapt to new data from either modality without compromising performance on previously learned tasks. Furthermore, the framework ensures robust segmentation capabilities even when only a single imaging modality (CT or PET) is provided. To evaluate the effectiveness of the proposed method, we conducted extensive experiments on a Head and Neck cancer segmentation and outcome prediction dataset (HECKTOR) \cite{hecktor_dataset}. The dataset contains data from seven different medical centers, where for each patient have we have a CT and PET scans, as well as EHR.

\section{Related Work}
\subsection{Segmentation Models}
Segmentation in medical imaging is essential for identifying anatomical structures, enabling precise diagnosis, treatment planning, and disease monitoring. One of the most widely used segmentation architectures is UNet \cite{unet}, which is designed with an encoder-decoder structure that enables localization while capturing contextual information through using skip connections. However, convolution-based architectures, such as UNet \cite{unet}, struggle to capture long-range dependencies. To address this, UNETR (UNet Transformer) was introduced \cite{unetr}. UNETR is a transformer-based model that reformulates 3D medical image segmentation as a sequence-to-sequence prediction task. Then, Swin UNETR \cite{swinunetr} built upon the UNETR framework by incorporating the Swin Transformer, which utilizes hierarchical self-attention and shifted windows to efficiently capture both local and global features, enhancing segmentation performance in medical imaging. Swin UNETR outperforms UNETR in PET/CT image segmentation tasks due to its ability to better capture cross-modal feature representations. For instance, in head and neck tumor segmentation, Swin UNETR achieved higher DSC scores compared to UNETR, demonstrating its superiority in handling multi-modal data \cite{swincross}. Studies have demonstrated that Swin UNETR exhibits strong generalizability across different centers and imaging conditions \cite{tta_swinunet}. However, its computational requirements are higher than those of UNETR due to the complexity of the Swin Transformer architecture, with some advancements introduced to increase its efficiency \cite{swinunetrv2}. 
\subsection{Existing Fusion Approaches}
Fusing different modalities in healthcare data, such as PET and CT scans, provides complementary information that enhances diagnostic accuracy disease characterization \cite{hybridatt_fusion, deep_ctpet, ex_petct, fuse_petct1}. From a medical imaging perspective, CT scans offer high-resolution anatomical details, while PET scans provide functional and metabolic insights. The fusion of the these two modalities allows a precise localization of abnormalities, such as tumors or inflammation \cite{breakthroughs_petct}. From a deep learning perspective, the integration of PET and CT scans enables models to take advantage of both spatial and functional features, enhancing the segmentation of the anatomical structure \cite{deep_ctpet}. Handling fusion of modalities in models can happen through early fusion, late fusion, or hybrid fusion \cite{multimodal_ai, fusion_healthcare, fusion_healthcare_2}. Early fusion combines images at the input level during feature extraction. This high-dimensional input increases memory and computational load and requires careful alignment of modalities. Late fusion processes each modality independently through feature extractors or models, then combines their outputs by averaging, computing the weighted sum, or learned fusion vectors. Separate models, in this case, increase inference time and lead to fewer interactions between modalities, alleviating the risk of missing cross-modal dependencies. On the other hand, hybrid fusion extracts intermediate features from separate modality-specific networks and fuses them at an intermediate layer before final decision-making. While this approach balances complexity, it requires careful design to identify the appropriate fusion points and account for the different learning dynamics between modalities \cite{fusion_healthcare, fusion_healthcare_2}. Thus, reaching an optimal fusion mechanism that efficiently learn from both modalities while reducing model complexity remains an open challenge. 
\subsection{Missing Modality}
The majority of multi-modal segmentation approaches rely on the existence of both modalities during model training. However, the availability of modalities for each patient is not always guaranteed in real-life scenarios. This is due to a variety of factors that are either clinical, technical, or because of cost constraints \cite{pet_challenges}. In some conditions, healthcare providers may find that CT only is necessary, or patients could have certain contraindications that requires them to avoid acquiring images of a certain modality \cite{limitations_ct_ctpet}. Moreover, failures in image acquisition can lead to missing modalities due to incomplete imaging data (e.g., motion artifacts, equipment malfunctions) \cite{ benefits_ctpet, pet_challenges}. Another reason is that PET scans are not always accessible in every healthcare facility, leading to missing PET images in resource-limited settings \cite{cost}. Lastly, collecting data from different hospitals and institutions raises the issue of inconsistent acquisitions and clinical practices, resulting in missing modalities in large medical datasets \cite{inconsistencies}. 

The aforementioned challenges require models to address this issue, leading to several research aspects attempting to tackle it. One approach is through data imputation and synthesis, where the missing modality is generated using the available one by using generative adversarial networks to learn the mapping between modalities \cite{alz_missing, diffusion_missing, synthesis_missing}. Another technique is to project multiple modalities into a shared latent space, capturing common features that can be utilized even when one modality is absent \cite{missing_survey}. This allows models to infer missing data during inference, maintaining performance despite incomplete inputs. Additionally, knowledge distillation is an approach that involves training a 'teacher' model on a complete multi-modal data and then guiding a 'student' model to perform effectively with missing modalities, leveraging learned knowledge to handle incomplete data scenarios \cite{missing_survey}. Thus, there is an increasing need to build architectures that are robust to missing modalities without compromising on the model performance on the modalities that include missing data \cite{missing_survey, missing_architecture}. 
\subsection{Parameter Efficient Fine-tuning (PEFT)}
Parameter-efficient fine-tuning (PEFT) is a technique in machine learning that allows us to adapt large pre-trained models to specific tasks without retraining the entire model \cite{peft_survey1}. This approach emerged as a response to the challenges posed by the increasing size of models, since fully retraining such large models for each new task is computationally expensive and resource-intensive. PEFT addresses this by modifying only the most relevant parameters, making the fine-tuning process more efficient while maintaining performance comparable to full fine-tuning \cite{peft_survey2, peft_survey1}. 

One of the commonly adopted PEFT techniques is low-rank adaptation (LoRA) \cite{lora} which freezes the pre-trained model's weights and injects trainable low-rank decomposition matrices into each layer of the Transformer architecture. This approach significantly reduces the number of trainable parameters for downstream tasks. For instance, when fine-tuning GPT-3 175B, LoRA can reduce the number of trainable parameters by 10,000 times and the GPU memory requirement by 3 times, while maintaining or even improving model performance compared to full fine-tuning \cite{lora}. Building upon LoRA, Weight-decomposed low-Rank adaptation (DoRA) \cite{dora} was proposed to further enhance fine-tuning efficiency and performance. DoRA decomposes the pre-trained weight into magnitude and direction for fine-tuning, specifically employing LoRA for directional updates to efficiently minimize the number of trainable parameters. This method enhances both the learning capacity and the training stability of LoRA while avoiding any additional inference overhead \cite{dora}. Additionally, another technique used is adding adapters to the model, which involves inserting small, trainable modules within each layer of a pre-trained model. During fine-tuning, only these adapter modules are updated, while the original model parameters remain unchanged \cite{adapters}.

In medical imaging, particularly for 3D image segmentation tasks, PEFT methods have gained prominence due to their ability to adapt large pre-trained models to specific tasks as annotated data are often limited, and computational efficiency is crucial. For example, large models such as the Segment Anything Model (SAM) \cite{sam} have shown impressive performance in natural image segmentation but often underperform in medical images due to domain differences. To bridge this gap, integrating LoRA into SAM has demonstrated improved performance in 3D medical image segmentation tasks while significantly reducing the number of trainable parameters \cite{lora_sam}. Moreover, prompt tuning -- a PEFT approach -- has been applied to models like UNet, enabling effective segmentation of CT images with minimal parameter updates \cite{prompt_tuning}. This approach has shown that prompt-tuned models can achieve performance close to fully fine-tuned models while adjusting only a small fraction of the parameters. Since medical imaging encompasses various modalities (e.g., CT, MRI), each with unique characteristics, PEFT methods have been developed to efficiently adapt models across these modalities. For example, the MA-SAM framework \cite{masam} incorporates 3D adapters into a 2D pre-trained model, allowing it to extract volumetric information from different medical imaging modalities. This adaptation has led to an improved segmentation performance in multiple data sets.

\begin{figure*}[t]
    \centering
    \includegraphics[width=1.0\linewidth]{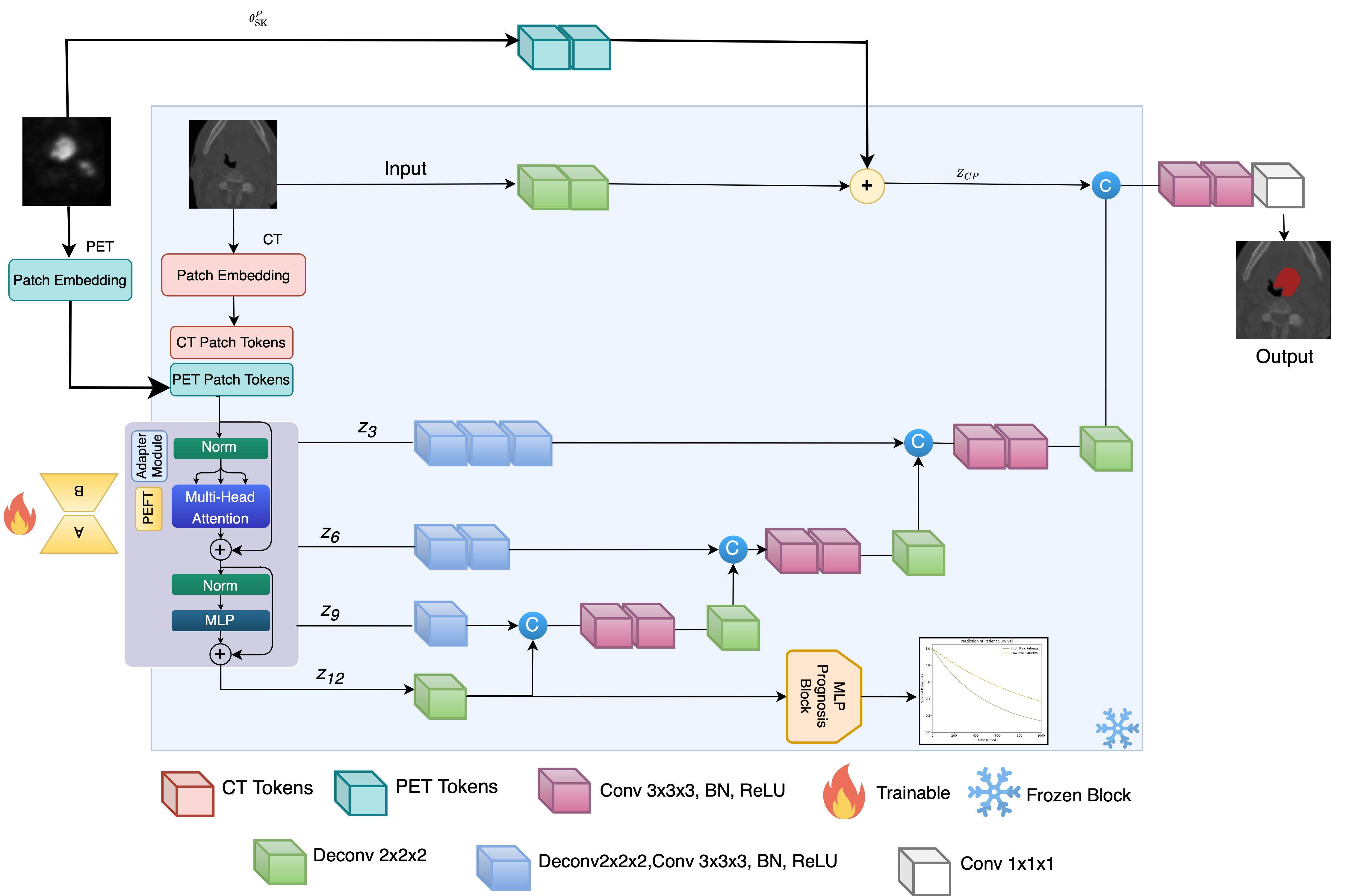}
    \caption{\textit{\textbf{Overview of our proposed architecture  PEMMA:}} At the input level, we separate the path for CT and PET by adding the PET Skip Connection $\theta_{\textrm{SK}}^{P}$. We freeze both the encoder and decoder part of the base segmentation model and introduce a PEFT module (LoRA or DoRA), after each ViT block (x12) as the only trainable layers. Additionally, our flexible architecture allows continual learning through adopting this model to other tasks, such as prognosis. 
    }
    \label{fig_main}
\end{figure*}

\subsection{Continual Learning}
Continual learning (CL) describes the model's ability to learn from a continuous stream of data, adapting to new information while retaining previously learned knowledge. It enables models to 1) improve over time without forgetting past experiences, 2) adjust to new tasks and environments seamlessly, and 3) reduce the need to retrain models from scratch, saving computational resources and time \cite{continual_survey}. Continuous adaptation and knowledge accumulation can lead to better decision-making and predictive accuracy and more effective interactions between humans and AI systems, including medical systems \cite{continual_medical}. Traditional approaches to CL include regularization-based methods, which constrain updates to prevent forgetting; rehearsal strategies, involving the replay of past data; and dynamic architectures that expand to accommodate new tasks \cite{continual_peft_context, continual_unified, continual_survey, continual_lora}. 

Integrating parameter-efficient fine-tuning techniques within CL frameworks has emerged as a promising strategy to enhance adaptability without incurring significant computational overhead \cite{continual_peft, continual_lora}. PEFT methods, such as LoRA, focus on adjusting a minimal subset of parameters, thereby reducing the risk of catastrophic forgetting and facilitating efficient knowledge transfer across tasks. In the domain of medical imaging, the application of CL combined with PEFT remains relatively unexplored. However, studies like \cite{continual_peft_missed} have demonstrated PEFT's effectiveness in adapting pre-trained models to medical imaging tasks, particularly in low-data regimes common in this field. These findings suggest that integrating PEFT into CL frameworks could enhance the adaptability and efficiency of models in medical imaging, though further research is needed to fully realize this potential.
 
\subsection{Our Previous Work}
 In our prior work presented at MICCAI 2023, we introduced PEMMA, a novel framework designed to efficiently adapt pre-trained transformer-based UNETR segmentation models \cite{unetr} from a single modality (CT) to a multi-modal setting (CT+PET). This adaptation uses minimal additional parameters \cite{pemma}. By leveraging Low-Rank Adaptation (LoRA) \cite{lora} and drawing inspiration from visual prompt tuning \cite{jia2022visual}, PEMMA strategically inserts learnable components, such as LoRA matrices and PET-specific skip connections, while keeping the original model parameters frozen. This approach enables flexible and lightweight integration of PET data, achieving segmentation performance that is comparable to or superior than that of early and late fusion methods, while using only 8\% of the trainable parameters. A key advantage of PEMMA is that it minimizes cross-modal entanglement, allowing the model to be updated using only one modality without suffering from catastrophic forgetting of the other modality. While PEMMA has demonstrated robustness across multi-center datasets and shown potential for continual learning scenarios, our current work builds upon this foundation by:

\begin{itemize}
    \item Modifying the Swin UNETR architecture to enable parameter-efficient multi-modal adaptation of a pre-trained model for new modalities and tasks.
    \item Further studying the generalization capabilities of PEMMA by employing the recently proposed PEFT method, DoRA, and conducting an extensive comparison.
    \item Investigating the task extensibility of PEMMA by extending the segmentation model to a new task: outcome prediction.
    \item Analyzing the qualitative performance of PEMMA on the segmentation task during both multi-modal adaptation and continual learning stages.
\end{itemize}

\section{Methodology}

\noindent \textbf{Problem Statement}: Let us consider a pre-trained transformer-based model for tumor segmentation, denoted by $F_{\Theta}^{C}: \mathcal{X}_C \rightarrow \mathcal{M}$, where $F_{\Theta}^{C}$ is the CT-only segmentation model with parameters $\Theta$, $\mathcal{X}_C$ represents the CT scan input space, and $\mathcal{M}$ denotes the output segmentation mask space. The main objective is to extend this model into a multi-modal framework $F_{\Phi}^{CP}: \mathcal{X}_C \times \mathcal{X}_P \rightarrow \mathcal{M}$ that can leverage both CT and PET scans for improved segmentation performance. Here, $F_{\Phi}^{CP}$ is the combined CT+PET model with parameters $\Phi$, and $\mathcal{X}_P$ is the PET input space. 

To achieve this, two key requirements must be met: (i) \textit{parameter efficiency}, meaning the number of parameters in the adapted model $|\Phi|$ should remain close to that of the original uni-modal model $|\Theta|$; and (ii) \textit{minimal cross-modal entanglement}, ensuring the model can be fine-tuned later using only one modality (CT or PET) without leading to forgetting the other modality (i.e., avoiding catastrophic forgetting).

\subsection{Standard Adaptation Methods}
\textbf{Early Fusion.} One of the most common strategies for training a multi-modal CT+PET segmentation model is to treat the two imaging modalities as separate input channels and merge them at the input level. This creates a new multi-modal input space $\mathcal{X}_{CP}$, over which a new model $\hat{F}^{CP}_{\Phi} : \mathcal{X}_{CP} \rightarrow \mathcal{M}$ is trained using both modalities. When applied to transformer-based architectures (e.g., UNETR \cite{unetr}, Swin UNETR \cite{swinunetr}), this method requires two key architectural modifications. First, the original uni-modal patch embedding layer, which converts the input into a sequence of patch tokens, must be replaced with a new multi-modal embedding layer. Second, there should be architectural changes in the model that account for the change in the input dimension. For example, for UNETR, the skip connection between the input and the decoder must be modified to support the increased number of channels.

Formally, let $x_C \in \mathcal{X}_C$ denote the CT image and let $\mathcal{T}^C_0 = \mathcal{E}^{C}_{\theta_{\text{PE}}}(x_C)$ be the CT patch tokens produced by the patch embedding layer of the CT-only model $F^C_\Theta$. Given a PET image $x_P \in \mathcal{X}_P$, the combined input is constructed as $x_{CP} = [x_C || x_P] \in \mathcal{X}_{CP}$, where $||$ represents channel-wise concatenation. A new multi-modal embedding layer $\mathcal{E}^{CP}_{\theta_{\text{PE}}}$ is then trained to produce the CT-PET patch tokens as $\mathcal{T}^{CP}_0 = \mathcal{E}^{CP}_{\theta_{\text{PE}}}(x_{CP})$, which are passed to the transformer encoder.

Similarly, if $z_C = \mathcal{S}^{C}_{\theta_{\text{SK}}}(x_C)$ is the output of the skip connection in the CT-only model, it is replaced with a new skip connection $\mathcal{S}^{CP}_{\theta_{\text{SK}}}$ trained to generate $z_{CP} = \mathcal{S}^{CP}_{\theta_{\text{SK}}}(x_{CP})$. With these two architectural changes, the adapted model can be fully trained on CT+PET data to enable multi-modal segmentation.

A key advantage of the early fusion approach is its parameter efficiency, as it introduces only a small number of additional parameters to the patch embedding and input skip connection layers to accommodate the increased number of input channels. However, a significant limitation is the entanglement of features from the CT and PET modalities in the shared representation. This entanglement can lead to catastrophic forgetting when fine-tuning the model on data from only a single modality (e.g., CT), potentially degrading performance on the other modality (e.g., PET).

Since the architectural modifications in early fusion are relatively minor, most parameters in the multi-modal model can be initialized directly from the pre-trained uni-modal (CT-only) model. For the newly introduced PET-specific parameters, we explore three initialization strategies: 1) \textit{random initialization:} weights are initialized stochastically, 2) \textit{zero initialization:} weights are set to zero, 3) \textit{cross-modal initialization:} PET-specific parameters are initialized using the corresponding weights from the CT branch of the pre-trained model. Among these, cross-modal initialization offers the benefit of leveraging prior knowledge from the CT modality to guide the PET channel, thereby accelerating convergence and improving fusion effectiveness.

\textbf{Late Fusion.} An alternative approach for incorporating an additional modality, such as PET, is to train a separate segmentation model $\mathcal{F}_{\Psi}^{P} : \mathcal{X}_P \rightarrow \mathcal{M}$ dedicated to PET scans, where $\mathcal{F}_{\Psi}^{P}$ denotes the PET-only model with parameters $\Psi$. The predictions from this model are then combined with those from the CT-only model $\mathcal{F}_{\Theta}^{C}$ at the output level.

Let $M_C \in \mathcal{M}$ and $M_P \in \mathcal{M}$ represent the segmentation masks predicted by the CT and PET models, respectively. The final fused prediction $M_{CP}$ is computed as a weighted combination of the two masks:

\begin{equation}
M_{CP} = w_C M_C + (1 - w_C) M_P,
\end{equation}

where $w_C$ is a scalar weight assigned to the CT modality. This late fusion strategy provides high flexibility in scenarios where the availability of modalities may vary across cases or over time. Each modality can be processed independently, and the final decision can be tailored by adjusting the fusion weights. However, it leads to a two-fold increase in the number of parameters ($|\Phi| = |\Theta| + |\Psi|$) and may not provide optimal accuracy.







\subsection{Proposed Adaptation Method: PEMMA}



In light of the strengths and weaknesses of existing adaptation methods, we introduce a novel framework (see Fig. \ref{fig_main}) for lightweight adaptation of a uni-modal model into a multi-modal model leveraging the inherent modularity of the transformer architecture. Our proposed framework, referred to as parameter-efficient multi-modal adaptation (PEMMA), is inspired by the concepts of visual prompt tuning (VPT) \cite{jia2022visual} and parameter-efficient finetuning (PEFT) \cite{lora, dora}. The PEMMA framework has three core components. Firstly, we introduce the new PET modality as a set of visual prompts (or context tokens) to the uni-modal model simply by adding a new patch embedding layer. Secondly, instead of fine-tuning all the parameters of the transformer encoder, we focus only on some of the layers and fine-tune them, such as the attention layers through LoRA and DoRA matrices. Finally, we introduce adaptation methods according to the segmentation model architecture used, allowimg seamless integration of the two modalities with accountability for cases of missing modalities. In the following sections, we explain the two approaches introduced to handle the multi-modal scenario in a parameter-efficient and accurate manner. We adopt two of the most commonly used transformers for segmentation tasks: UNETR and Swin UNETR.

\subsubsection{UNETR}
We start by implementing the aforementioned framework through the UNETR framework. Given the PET image $\mathbf{x}_{P}$, we first add a new PET patch embedding layer $\mathcal{E}^{P}_{\theta_{\text{PE}}}$, which generates a new set of $N$ PET patch tokens as $\mathcal{T}_0^{P} = \mathcal{E}^{P}_{\theta_{\text{PE}}}(\mathbf{x}_{P})$ that are passed to the subsequent transformer blocks in the encoder. Now, the operations of a transformer block can be represented as $\{\mathcal{T}_{\ell}^{C},\mathcal{T}_{\ell}^{P}\} = \mathcal{G}_{\theta_{\ell}}(\{\mathcal{T}_{\ell-1}^{C},\mathcal{T}_{\ell-1}^{P}\})$. Note that the above modification increases the number of tokens processed by the transformer encoder from $N$ to $2N$. However, the decoder in the pre-trained uni-modal model can handle only $N$ tokens. In order to avoid making changes to the decoder architecture, we allow only the $N$ CT tokens from the intermediate transformer blocks ($\mathcal{T}_{\ell}^{C}$) to pass through to the decoder. Refer to Table \ref{UNETR_ablation} for Dimensionality Reduction ablation study. It must be emphasized that the self-attention architecture ensures that the knowledge from the PET tokens gets distilled into the CT tokens, even though the PET tokens are ignored by the decoder layers. This is similar to VPT, where the learnable prompts only serve as the context and this contextual information gets distilled into the other tokens, even when the additional prompts are ignored in the end.

The multi-head self-attention parameters of a transformer block $\ell$ can be considered as a collection of four weight matrices denoted as $\{\mathbf{W}_{O,\ell},\mathbf{W}_{Q,\ell},\mathbf{W}_{K,\ell},\mathbf{W}_{V,\ell}\}$. In LoRA, the updates to $\mathbf{W}_{Q,\ell}$ and $\mathbf{W}_{V,\ell}$ are decomposed into a pair of low rank matrices $\mathbf{A} \in \mathbb{R}^{r \times d}$ and $\mathbf{B} \in \mathbb{R}^{d \times r}$, where $r$ represents the rank of the two matrices. Let $h_{*,\ell}$ and $\tilde{h}_{*,\ell}$ be the input and output, respectively, of an attention layer in the $\ell^{\text{th}}$ block. Then, LoRA operation can be summarized as: 

\begin{equation}
    \label{eq:processing_lora}
    \begin{split}
        \tilde{h}_{Q, \ell} &= \mathbf{W}_{Q,\ell}  h_{Q, \ell} + \alpha \mathbf{B}_{Q,\ell}  \mathbf{A}_{Q,\ell}   h_{Q, \ell}, \\ 
         \tilde{h}_{V, \ell} &= \mathbf{W}_{V,\ell}  h_{V, \ell} + \alpha  \mathbf{B}_{V,\ell}  \mathbf{A}_{V,\ell}   h_{V, \ell},
    \end{split}
\end{equation} 

\noindent where $\alpha$ is a fixed scalar. Note that parameter-efficiency is achieved by allowing $\theta_{\textrm{LoRA}} = \{\mathbf{A}_{Q,\ell},\mathbf{A}_{V,\ell},\mathbf{B}_{Q,\ell},\mathbf{B}_{V,\ell}\}_{\ell=1}^L$ as the only learnable parameters and freezing the rest of the parameters in the transformer encoder. 

Finally, in contrast to the early fusion approach where $\mathcal{S}_{\theta_{\textrm{SK}}^{C}}$ is replaced with $\mathcal{S}_{\theta_{\textrm{SK}}^{CP}}$, we leave  $\mathcal{S}_{\theta_{\textrm{SK}}^{C}}$ untouched and introduce an additional parallel path $\mathcal{S}_{\theta_{\textrm{SK}}^{P}}$ for the PET image. Let $\mathbf{z}_{P} = \mathcal{S}_{\theta_{\textrm{SK}}^{P}}(\mathbf{x}_P)$ be the output of the additional direct skip connection layer introduced for the PET modality. In this case, the combined output of the input skip layer is $\mathbf{z}_{CP} = \mathbf{z}_{C} + \beta \mathbf{z}_{P}$, where $\beta$ is the weight assigned to the PET modality. The main motivation for the introduction of new patch embedding ( $\mathcal{E}_{\theta_{\textrm{PE}}^{P}}$) and input skip layers ($\mathcal{S}_{\theta_{\textrm{SK}}^{P}}$) for the PET modality (rather than replacing them as in early fusion) is to minimize cross-modal entanglement. With the introduction of these additional layers, it is possible to subsequently fine-tune the multi-modal model using only one of the modalities, without affecting the model's ability to handle the other modality. 

\begin{figure*}
    \centering
    \includegraphics[width=1.0\linewidth]{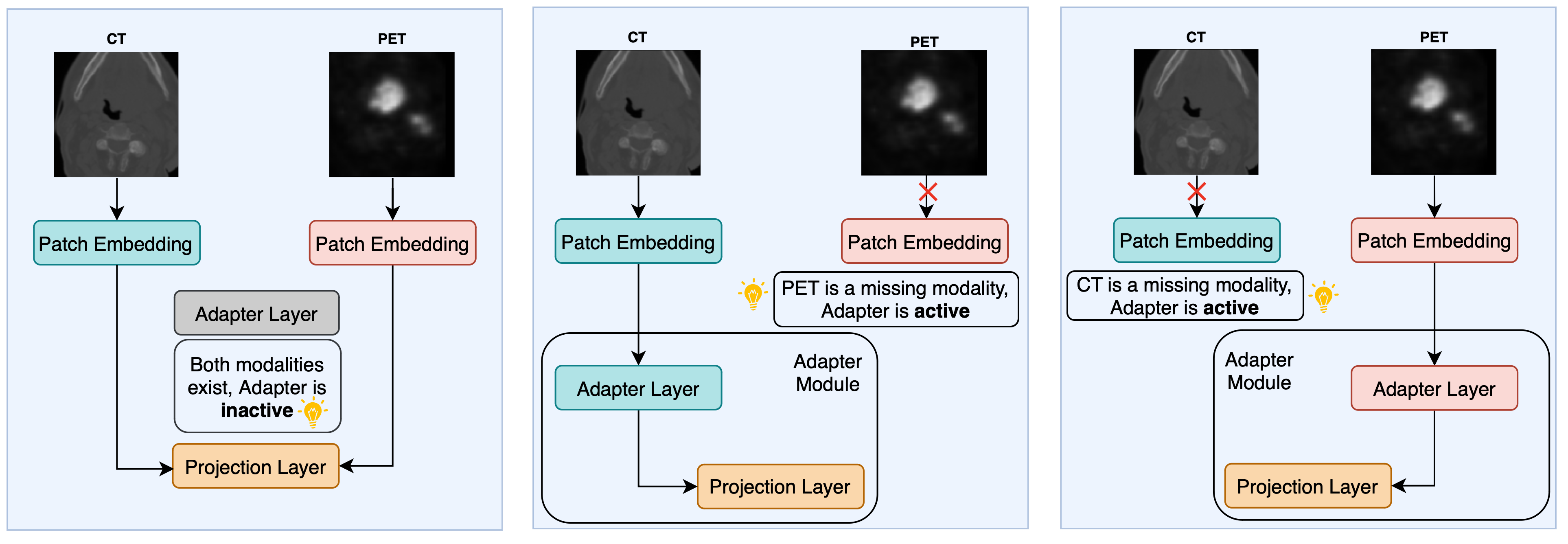}
    \caption{\textit{\textbf{Adapter Module:} } At the input level, we pass the CT and PET images to a patch embedding layer. The adapter module includes an adapter layer and a projection layer. When both modalities exist (\textit{left}), the adapter is inactive and the patch tokens pass to the projection layer. When either modality is missing (\textit{middle}/\textit{right}), the adapter layer gets activated and the patch tokens pass through the adapter layer followed by the projection layer. The adapter layer reshapes the single-modality input to the expected multi-modality input required by the model's architecture.
    }
    \label{fig:adapters}
\end{figure*}

\subsubsection{Swin UNETR}
Adding to the previous work done in PEMMA that used the UNETR architecture, we extend it to include Swin UNETR. Swin UNETR leverages the Swin Transformer as its encoder, which introduces shifted window-based attention to compute self-attention locally within non-overlapping windows. This approach significantly reduces the computational cost compared to global self-attention used in standard Vision Transformers, making it more efficient and scalable for high-resolution 3D medical images. The hierarchical structure of Swin Transformers also facilitates the extraction of multi-scale features, which aligns well with the UNet-style decoder and contributes to improved segmentation performance.

Despite these advantages, Swin UNETR presents certain challenges. Its architecture is inherently more complex due to the use of window partitioning, hierarchical stages, and localized attention mechanisms. This added complexity requires more careful design considerations when adapting the model to handle additional modalities such as PET and EHR, especially in terms of input alignment, patch embedding, and skip connections. In the case of UNETR, the sequence of flattened patch embeddings from the input volume of CT and PET scans is passed to the transformer encoder with a few changes in the multi-modal model as compared to the uni-modal one. However, Swin UNETR requires passing a grid of 3D patch embeddings produced by a Conv3D layer to the windowed self-attention layer, which is not directly possible when using a multi-modal model as it is challenging to change the input dimension of Swin UNETR from the uni-modal case. Additionally, handling cases of missing modalities remains challenging, necessitating a dynamic way of dealing with the existence of one or more modalities. Thus, this requires architectural changes, leading to introducing the adapter layer which allows passing the multi-modal representation to Swin UNETR in a way that also accounts for cases of missing modalities. 


Similarly to UNETR, given the PET image $\mathbf{x}_{P}$, we create the PET patch embedding layer $\mathcal{E}^{P}_{\theta_{\text{PE}}}$, which generates a new set of $N$ PET patch tokens as $\mathcal{T}_0^{P} = \mathcal{E}^{P}_{\theta_{\text{PE}}}(\mathbf{x}_{P})$ that are passed to subsequent transformer blocks in the encoder. Then, these patch embeddings cannot be directly passed to the transformer block, which is represented as $\{\mathcal{T}_{\ell}^{C},\mathcal{T}_{\ell}^{P}\} = \mathcal{G}_{\theta_{\ell}}(\{\mathcal{T}_{\ell-1}^{C},\mathcal{T}_{\ell-1}^{P}\})$. Due to this, we introduce the \textit{adapter layer} with parameters $\theta_\textrm{Adapter}$ that is used to transform the input into a feasible format for the windowing attention module. This adapter simulates the presence of both modalities by reshaping the single-modality input into the expected two-channel format. The adapted input is then passed to a projection layer before going to the standard Swin UNETR transformer block. When both modalities are present, CT and PET volumes are concatenated along the channel dimension and passed directly to the projection layer without adapting it first. This process is further clarified in Figure \ref{fig:adapters}. In this way, the adapter module 1) overcomes the challenge of passing the multi-modal input to the windowing attention module and 2) handles the cases of missing modalities by activating the adaptation layer when only one modality is available.

To encourage generalization and robustness, we apply a \textit{modality dropout schedule} during training, where some of batches include both CT and PET, while the others include only CT or only PET. This ensures that the model learns to operate effectively under different missing modality conditions.

For parameter-efficient adaptation, we incorporate LoRA  \cite{lora} (as used in UNETR) and its extension \textbf{DoRA} (Decomposed Rank Adaptation \cite{dora}). DoRA performs low-rank adaptation on the \textit{direction} of attention weights while preserving their \textit{magnitude}. Let $\mathcal{L}$ denote the number of transformer blocks, and let $\theta_{\text{DoRA}} = \{A_{\ast,\ell}, B_{\ast,\ell}\}_{\ell=1}^{\mathcal{L}}$ represent the set of learnable low-rank parameters. For a given attention projection matrix $W_{\ast,\ell}$, DoRA performs the update:

\[
\tilde{W}_{\ast,\ell} = \text{normalize}(W_{\ast,\ell}) + \alpha B_{\ast,\ell} A_{\ast,\ell}
\]

where $\alpha$ is a scaling factor, and the normalization ensures the original weight's magnitude is preserved. This enables efficient, stable, and directionally-informed fine-tuning across tasks and modalities with a minimal number of learnable parameters.

\begin{table*}
\scriptsize	
\caption{Distribution of the HECKTOR dataset, acquired from seven different medical centers, indicating the different centers' data used at each stage.}
\begin{tabular}{m{2cm} m{3.5cm} m{2cm} c m{1.3cm} m{2cm}}
\hline
\textbf{Stages} & \textbf{Center} & \textbf{City, Country} & \textbf{Acronym} & \textbf{Number of Samples} & \textbf{Pet/CT Scanner} \\ \hline
\multirow{4}{*}{\textbf{Pre-training}} &
 Centre hospitalier universitaire de Sherbooke, Sherbrooke, CA &
 Sherbrooke, Canada &
 CHUS & \centering
 72 &
 GeminiGXL 16, Philips, Siemens \\ \cline{2-6}
 &
 Centre Hospitalier Universitaire Vaudois, CH &
 Vaud, Switzerland &
 CHUV & \centering
 53 &
 Discovery D690 TOF, GE Healthcare \\ \cline{2-6}
 &
 Centre hospitalier de l’Université de Montréal, Montréal, CA &
 Montreal, Canada &
 CHUM & \centering
 56 &
 Discovery STE, GE Healthcare \\ \cline{2-6}
 &
 Centre Hospitalier Universitaire de Poitiers, FR &
 Poitiers, France &
 CHUP & \centering
 72 &
 Biograph mCT 40 ToF, Siemens \\ \hline
\textbf{Multi-Modal Adaptation} &
 MD Anderson Cancer Center, Houston, Texas, USA &
 Texas, USA &
 MDA & \centering
 197 &
 Discovery HR, RX, ST, and STE (GE Healthcare) \\ \hline
\textbf{CL Task 1} &
 Hôpital général juif, Montréal, CA &
 Montreal, Canada &
 HGJ & \centering
 55 &
 Discovery ST, GE Healthcare \\ \hline
\textbf{CL Task 2} &
 Hôpital Maisonneuve-Rosemont, Montréal, CA &
 Montreal, Canada &
 HMR & \centering
 18 &
 Discovery ST, GE Healthcare \\ \hline
\end{tabular}
\label{data_dist}
\end{table*}

\begin{table} [!h]
\caption{Clinical characteristics of patients in the dataset (apart from age and weight).}
\label{table:clinical_characteristics_revised_2}
\centering
\begin{tabular}{lc}
\toprule
{\textbf{Characteristic}} & {\textbf{Dataset (n = 488)}} \\
\midrule
\textbf{Demographics} \\
\quad Gender: \\
\quad \quad Male & 402 (82.4\%) \\
\quad \quad Female & 86 (17.6\%) \\
\midrule
\textbf{Lifestyle Factors} \\
\quad Alcohol Consumption: \\
\quad \quad Yes & 95 (19.5\%) \\
\quad \quad No & 59 (12.1\%) \\
\quad \quad Unknown & 334 (68.4\%) \\
\quad Tobacco Consumption: \\
\quad \quad Yes & 85 (17.4\%) \\
\quad \quad No & 105 (21.5\%) \\
\quad \quad Unknown & 298 (61.1\%) \\
\midrule
\textbf{Disease-Related Factors} \\
\quad HPV Status: \\
\quad \quad Positive & 274 (56.1\%) \\
\quad \quad Negative & 43 (8.8\%) \\
\quad \quad Unknown & 171 (35.1\%) \\
\quad Performance Status: \\
\quad \quad 0 & 86 (17.7\%) \\
\quad \quad 1 & 114 (23.3\%) \\
\quad \quad 2 & 11 (2.3\%) \\
\quad \quad 3 & 3 (0.6\%) \\
\quad \quad 4 & 1 (0.2\%) \\
\quad \quad Unknown & 273 (55.9\%) \\
\quad RFS: \\
\quad \quad Uncensored & 96 (19.7\%) \\
\quad \quad Censored & 392 (80.3\%) \\
\midrule
\textbf{Treatment-Related Factors} \\
\quad Surgery: \\
\quad \quad Yes & 50 (10.3\%) \\
\quad \quad No & 248 (50.8\%) \\
\quad \quad Unknown & 190 (38.9\%) \\
\quad Chemotherapy: \\
\quad \quad Yes & 422 (86.5\%) \\
\quad \quad No & 66 (13.5\%) \\
\bottomrule
\multicolumn{2}{p{8cm}}{Note: Unknown values may reflect incomplete EHR data or patient refusal to disclose information.} \\
\end{tabular} \label{EHR_data}
\end{table}

\subsubsection{Flexible Training and Inference Strategy}
The PEMMA framework introduces three new parameters $\theta_{\textrm{PE}}^{P}$, $\theta_{\textrm{PEFT}}$, $\theta_{\textrm{SK}}^{P}$ that work for both segmentation architectures used, along with $\theta_{\textrm{Adapter}}$ in case of Swin UNETR (we consider $\theta_{\textrm{PEFT}}$ as either $\theta_{\textrm{LoRA}}$ or $\theta_{\textrm{DoRA}}$). When adapting the uni-modal model to the multi-modal scenario, both CT and PET training data is required and all the new parameters $\{\theta_{\textrm{PE}}^{P},\theta_{\textrm{PEFT}},\theta_{\textrm{SK}}^{P}, \theta_{\textrm{Adapter}}\}$ are learned, while the parameters of the pre-trained unimodal model $\Theta$ are completely frozen. Thus, the parameters of the multi-modal model are $\Phi = \{\Theta,\theta_{\textrm{PE}}^{P},\theta_{\textrm{PEFT}},\theta_{\textrm{SK}}^{P}, \theta_{\textrm{Adapter}}\}$ and $|\Phi|$ is only marginally higher than $|\Theta|$ (hence, parameter-efficient). Subsequently, if new data is available to update the multi-modal model, only $\theta_{\textrm{PEFT}}$ needs to be updated and all other parameters can be frozen. This allows the flexibility of fine-tuning the multi-modal model using one or both modalities. Similarly, the multi-modal allows flexible inference - it can effectively utilize both modalities when they are available, but can also be applied to only a single modality (albeit with some degradation in the segmentation accuracy). 

Moreover, if our model is to be adopted by a data center that has a different data distribution in their CT and PET scans, our architectural design of Swin UNETR enables finetuning the adapter layer independently to consider the distribution shift. This continual learning technique provides an efficient way for users to use our model even when only a single modality is available at inference time with much less computational needs as compared to re-training from scratch in case of distribution shifts. 

Despite that the model was mainly designed for segmentation, our flexible framework allows users to continually perform other tasks, such as prognosis. In case of prognosis, users can adopt the encoder part of our model, while still keeping the advantages of training through a flexible multi-modal approach and in a parameter-efficient matter, as well as effectively handling the missing modalities. 


\section{Experimental Setting}
\subsection{Dataset}
The dataset used in this study is the \textbf{HEad and neCK TumOR (HECKTOR)} dataset, publicly released on the MICCAI 2022 challenge website \cite{hecktor_dataset}. It is a multi-center, multi-class, and multi-modal dataset comprising data from 523 patients collected across seven medical centers. These patients were diagnosed with histologically proven oropharyngeal head and neck cancer and underwent radiotherapy and/or chemotherapy treatment planning. A detailed breakdown of the patient distribution across the centers and the scanner types used for acquisition is provided in Table \ref{data_dist}.

For each patient, the dataset includes:
1) CT and PET scans: Imaging data capturing the anatomical and metabolic characteristics of the tumors; 2) Segmentation masks: Outlines of the primary gross tumor volumes (GTVp) and nodal gross tumor volumes (GTVn), providing ground truth annotations for the scans; 3) Electronic health records (EHR): Clinical information, including Recurrence-Free Survival (RFS) data such as time-to-event and censoring status. Approximately 70\% of the patients in the dataset are censored. Table \ref{EHR_data} presents the clinical characteristics of the 488 patients for whom survival label data was available. The table summarizes the distribution of several key variables extracted from EHR data, such as gender, alcohol and tobacco consumption, HPV status, performance status, surgery, and chemotherapy. The number of patients and the corresponding percentage are provided for each characteristic. This data was employed in the prognosis experiments.

The PET volumes are registered to CT volumes at a common origin, though they vary in size and resolution: 1) CT scans size ranges from (128, 128, 67) to (512, 512, 736) voxels, while its resolution ranges from (0.488, 0.488, 1.00) to (2.73, 2.73, 2.80) mm in the x, y, and z directions. On the other hand, PET scans' size ranges from (128, 128, 66) to (256, 256, 543) voxels, while its resolution ranges from (2.73, 2.73, 2.00) to (5.47, 5.47, 5.00) mm in the same directions. 

The coverage of the scans differs across patients: some are limited to the head and neck regions, while others encompass the full body, extending up to the abdomen. All CT and PET scans are provided in the NIFTI format. As for the segmentation masks, they were annotated by medical professionals at the respective centers, ensuring clinical accuracy and reliability.

\subsection{Implementation Details}

The PyTorch framework \cite{pytorch} was used for implementing the proposed method.

\subsubsection{Optimization}
\textit{Segmentation}: Both UNETR and Swin UNETR models were trained using the AdamW optimizer with a cosine annealing scheduler. A weight decay of 1e-5 was used. The initial learning rate varied across training phases: 1e-3 for pre-training, 1e-4 for multi-modal adaptation, and 1e-4 for continual learning experiments. Dice and Cross Entropy loss were employed as loss functions. Early stopping, with a patience of 20 epochs, was also implemented. Training was conducted on four Nvidia A6000 RTX 48GB GPUs, using a global batch size of 8. The models were trained for 300 epochs during the pre-training phase, 250 epochs for the multi-modal adaptation phase, and 50 epochs for both continual learning experiments. 

\textit{Prognosis}: The prognosis training implementation is similar to segmentation, except that it was performed using only the encoder of Swin UNETR with a prognosis head that consists of two MLP layers. Training was conducted on Nvidia 5000 RTX 32GB GPU, using a batch size of 2. The initial learning rate is 1e-3 for multi-modal adaptation and continual learning using CT and CT+PET, and 1e-4 for multi-modal adaptation and continual learning using CT+PET with EHR. The weight decay is 1e-5. DeepHit \cite{deephit} loss is implemented with 20 discrete time bins corresponding to the quantiles of the survival time distribution for a total of 150 epochs, with an early stopping patience of 20 epochs.

\subsubsection{Augmentation}
Data augmentation techniques are employed using the MONAI \cite{monai} transforms to enhance the diversity and generalization of our training dataset, which consists of registered CT and PET scans for each patient. A uniform sampling strategy ensures equal selection probability across datasets. The preprocessing pipeline begins by loading the images and ensuring a channel-first format using \monaifunc{LoadImage}. Both CT and PET volumes are then resampled to an isotropic spacing of $1.0 \times 1.0 \times 1.0\,\text{mm}^3$. Intensity processing is performed modality-specifically: CT intensities are clipped and normalized to the $[0, 1]$ range using \monaifunc{ScaleIntensityRange} with arguments \monaiarg{a\_min=-200} and \monaiarg{a\_max=250}, followed by stochastic intensity augmentations including \monaifunc{RandScaleIntensity} (\monaiarg{factors=$\pm$0.1}, \monaiarg{prob=0.5}), \monaifunc{RandShiftIntensity} (\monaiarg{offsets=$\pm$0.1}, \monaiarg{prob=0.5}), and \monaifunc{RandGaussianNoise} (\monaiarg{std=0.01}, \monaiarg{prob=0.2}); PET intensities are normalized based on non-zero values using \monaifunc{NormalizeIntensity}. Subsequently, the \monaifunc{Orientation} of the combined image and the corresponding label map is standardized to "RAS". To focus training on relevant areas, \monaifunc{RandCropByPosNegLabel} extracts random patches of size $96 \times 96 \times 96$, maintaining a positive (foreground) to negative (background) sample ratio of 2:1 based on the label map. Finally, spatial augmentations are applied to these cropped patches: \monaifunc{RandAffine} (\monaiarg{prob=0.5}) introduces slight rotations ($\pm 9^{\circ}$ per axis) and scaling ($\pm 10\%$ per axis) using bilinear interpolation for the image and nearest-neighbor for the label with zero padding; \monaifunc{RandFlip} (\monaiarg{prob=0.2}) performs random flips along spatial axes; and \monaifunc{RandRotate90} (\monaiarg{prob=0.2}) applies random 90-degree rotations.

\subsubsection{Network Structures}
Our proposed network architecture adopts the UNETR \cite{unetr} and Swin UNETR models~\cite{swinunetr} for multi-modal 3D medical image segmentation, specifically designed to process registered CT and PET scans. The core encoder in UNETR and Swin UNETR leverages a modified \monaifunc{Transformer} and \monaifunc{SwinTransformer} module respectively, tailored for flexible handling of input modalities specified by the \monaiarg{mode} parameter (\texttt{'ct'}, \texttt{'pet'}, or \texttt{'ctpet'}). The selection of different modes during training for robust learning and generalizability was set to  (\texttt{'ct' = 0.2}, \texttt{'pet' = 0.2}, or \texttt{'ctpet' 0.6}) .  Within these transformers, distinct patch embedding layers are instantiated for CT and PET, each processing its corresponding modality and projecting to \monaiarg{embed\_dim}. For combined CT+PET processing (\monaiarg{mode='ctpet'}), the outputs of both embeddings are concatenated and processed through the transformer block hierarchy consisting of multiple stages. The outputs at four different stages interface with a UNet style decoder. The decoder path consists of modules performing upsampling and feature concatenation via skip connections.

\begin{table*}
 \centering
 \caption{Performance comparison of Multi-modal Adaptation (MDA) strategies applied to UNETR ($\Phi$=92.58M total parameters) and Swin UNETR ($\Psi$=62.19M total parameters) backbones, initially pre-trained on CT-only data. Adaptation was performed using combined CT+PET (CP) training data. The table shows Tumor, Lymph Node, and Average Dice scores achieved during inference on the adaptation dataset distribution (MDA) and two new datasets (HGJ, HMR). Evaluations were conducted using both CT and PET (CP), CT-only, and PET-only inputs. We compare the efficiency of each approach based on the number of additional trainable parameters relative to $\Phi$ or $\Psi$) required for adaptation. CP=CT+PET.}
 \resizebox{\textwidth}{!}{%
   \setlength{\tabcolsep}{0.5em} 
   {\renewcommand{\arraystretch}{1.5}
     \begin{tabular}{@{}c|ccccc|cccccc|cccccc@{}}
       \toprule
       \rowcolor[HTML]{C0C0C0}
       \textbf{Models} & \textbf{Dataset} & \multicolumn{1}{c}{\cellcolor[HTML]{C0C0C0}$\longrightarrow$} & \multicolumn{3}{c|}{\cellcolor[HTML]{C0C0C0}\textbf{Multi-modal Adaptation (MDA)}} & \multicolumn{6}{|c|}{\cellcolor[HTML]{C0C0C0}\textbf{New Dataset (HGJ)}} & \multicolumn{6}{c}{\cellcolor[HTML]{C0C0C0}\textbf{New Dataset (HMR)}} \\
       & \textbf{Train Modalities} & \multicolumn{1}{c}{$\longrightarrow$} & \multicolumn{3}{c|}{\textbf{CP}} & \multicolumn{3}{c|}{\textbf{CT}} & \multicolumn{3}{c|}{\textbf{CP}} & \multicolumn{3}{c|}{\textbf{CT}} & \multicolumn{3}{c}{\textbf{CP}} \\
       \rowcolor[HTML]{C0C0C0}
       & \textbf{Infer Modalities} & \multicolumn{1}{c}{\cellcolor[HTML]{C0C0C0}$\longrightarrow$} & \multicolumn{1}{c}{\cellcolor[HTML]{C0C0C0}\textbf{CP}} & \multicolumn{1}{c}{\cellcolor[HTML]{C0C0C0}\textbf{CT}} & \multicolumn{1}{c|}{\cellcolor[HTML]{C0C0C0}\textbf{PET}} & \multicolumn{1}{c}{\cellcolor[HTML]{C0C0C0}\textbf{CP}} & \multicolumn{1}{c}{\cellcolor[HTML]{C0C0C0}\textbf{CT}} & \multicolumn{1}{c|}{\cellcolor[HTML]{C0C0C0}\textbf{PET}} & \multicolumn{1}{c}{\cellcolor[HTML]{C0C0C0}\textbf{CP}} & \multicolumn{1}{c}{\cellcolor[HTML]{C0C0C0}\textbf{CT}} & \multicolumn{1}{c|}{\cellcolor[HTML]{C0C0C0}\textbf{PET}} & \multicolumn{1}{c}{\cellcolor[HTML]{C0C0C0}\textbf{CP}} & \multicolumn{1}{c}{\cellcolor[HTML]{C0C0C0}\textbf{CT}} & \multicolumn{1}{c|}{\cellcolor[HTML]{C0C0C0}\textbf{PET}} & \multicolumn{1}{l}{\cellcolor[HTML]{C0C0C0}\textbf{CP}} & \multicolumn{1}{c}{\cellcolor[HTML]{C0C0C0}\textbf{CT}} & \multicolumn{1}{c}{\cellcolor[HTML]{C0C0C0}\textbf{PET}} \\
       \midrule
       \multirow{9}{*}{\textbf{UNETR}} & \multirow{3}{*}{\begin{tabular}[c]{@{}c@{}}\textbf{Late Fusion}\\ \textcolor{red}{(\textbf{\large params=2} $\Phi$)}\end{tabular}} & \large Tumor & \large0.69 & \large0.37 & \large0.43 & \large0.67 & \large0.45 & \multicolumn{1}{c|}{\large0.32} & \large0.40 & \large0.36 & \large0.23 & \large0.68 & \large0.43 & \multicolumn{1}{c|}{\large0.05} & \large0.47 & \large0.46 & \large0.24 \\
       & & \large Lymph & \large0.64 & \large0.51 & \large0.51 & \large0.61 & \large0.56 & \multicolumn{1}{c|}{\large0.21} & \large0.48 & \large0.46 & \large0.27 & \large0.56 & \large0.52 & \multicolumn{1}{c|}{\large0.32} & \large0.30 & \large0.34 & \large0.11 \\
       & & \large Avg & \large0.67 & \large0.44 & \large0.47 & \large0.64 & \large0.51 & \multicolumn{1}{c|}{\large0.27} & \large0.44 & \large0.41 & \large0.25 & \large0.62 & \large0.48 & \multicolumn{1}{c|}{\large0.19} & \large0.39 & \large0.40 & \large0.18 \\
       \cmidrule{2-18}
       & \multirow{3}{*}{\begin{tabular}[c]{@{}c@{}} \textbf{Early Fusion}\\  \textcolor{red}{(\textbf{\large params=1.0043} $\Phi$)}\end{tabular}} & \large Tumor & \large0.65 & \large0.41 & \large0.68 & \large0.76 & \large0.71 & \multicolumn{1}{c|}{\large0.17} & \large0.81 & \large0.42 & \large0.30 & \large0.82 & \large0.34 & \multicolumn{1}{c|}{\large0.01} & \large0.63 & \large0.35 & \large0.42 \\
       & & \large Lymph & \large0.63 & \large0.48 & \large0.64 & \large0.63 & \large0.64 & \multicolumn{1}{c|}{\large0.30} & \large0.56 & \large0.48 & \large0.40 & \large0.58 & \large0.38 & \multicolumn{1}{c|}{\large0.34} & \large0.56 & \large0.35 & \large0.25 \\
       & & \large Avg & \large0.64 & \large0.45 & \large0.66 & \large0.70 & \large0.68 & \multicolumn{1}{c|}{\large0.24} & \large0.69 & \large0.45 & \large0.34 & \large0.70 & \large0.36 & \multicolumn{1}{c|}{\large0.03} & \large0.60 & \large0.35 & \large0.34 \\
       \cmidrule{2-18}
       & \multirow{3}{*}{\begin{tabular}[c]{@{}c@{}} \textbf{PEMMA (LoRA)}\\ \textcolor{red}{(\textbf{\large params=0.08} $\Phi$)} \end{tabular}} & \large Tumor & \large0.67 & \large0.41 & \large0.64 & \large0.82 & \large0.70 & \multicolumn{1}{c|}{\large0.29} & \large0.81 & \large0.43 & \large0.31 & \large0.86 & \large0.38 & \multicolumn{1}{c|}{\large0.34} & \large0.64 & \large0.41 & \large0.39 \\
       & & \large Lymph & \large0.60 & \large0.50 & \large0.63 & \large0.61 & \large0.72 & \multicolumn{1}{c|}{\large0.30} & \large0.57 & \large0.48 & \large0.40 & \large0.68 & \large0.50 & \multicolumn{1}{c|}{\large0.26} & \large0.57 & \large0.48 & \large0.26 \\
       & & \large Avg & \large0.64 & \large0.45 & \large 0.64 & \large0.72 & \large0.71 & \multicolumn{1}{c|}{\large0.30} & \large0.72 & \large0.45 & \large0.33 & \large0.75 & \large0.48 & \multicolumn{1}{c|}{\large0.34} & \large0.63 & \large0.42 & \large0.33 \\
        \midrule
       \multirow{9}{*}{\textbf{Swin UNETR}} & \multirow{3}{*}{\begin{tabular}[c]{@{}c@{}} \textbf{Early Fusion}\\  \textcolor{red}{(\textbf{\large params=1.0043} $\Psi$)}\end{tabular}} & \large Tumor & \large 0.80 & \large 0.65 & \large 0.66 & \large 0.77 & \large 0.79 & \multicolumn{1}{c|}{\large 0.41} & \large 0.80 & \large 0.77 & \large 0.70 & \large 0.70 & \large 0.62 & \multicolumn{1}{c|}{\large 0.60} & \large 0.68 & \large 0.62 & \large 0.50 \\
       & & \large Lymph & \large 0.72 & \large 0.71 & \large 0.44 & \large 0.69 & \large 0.63 & \multicolumn{1}{c|}{\large 0.65} & \large 0.74 & \large 0.67 & \large 0.44 & \large 0.58 & \large 0.48 & \multicolumn{1}{c|}{\large 0.24} & \large 0.56 & \large 0.30 & \large 0.40 \\
       & & \large Avg & \large 0.76 & \large 0.68 & \large 0.55 & \large 0.73 & \large 0.71 & \multicolumn{1}{c|}{\large 0.53} & \large 0.77 & \large 0.72 & \large 0.57 & \large 0.64 & \large 0.55 & \multicolumn{1}{c|}{\large 0.42} & \large 0.62 & \large 0.46 & \large 0.45 \\
       \cmidrule{2-18}
       & \multirow{3}{*}{\begin{tabular}[c]{@{}c@{}} \textbf{PEMMA (LoRA)}\\ \textcolor{red}{(\textbf{\large params=0.005} $\Psi$)} \end{tabular}} & \large Tumor & \large 0.76 & \large 0.66 & \large 0.64 & \large 0.81 & \large 0.81 & \multicolumn{1}{c|}{\large 0.65} & \large 0.82 & \large 0.80 & \large 0.75 & \large 0.74 & \large 0.66 & \multicolumn{1}{c|}{\large 0.66} & \large 0.70 & \large 0.64 & \large 0.53 \\
       & & \large Lymph & \large 0.75 & \large 0.70 & \large 0.47 & \large 0.72 & \large 0.70 & \multicolumn{1}{c|}{\large 0.39} & \large 0.73 & \large 0.69 & \large 0.42 & \large 0.63 & \large 0.55 & \multicolumn{1}{c|}{\large 0.34} & \large 0.60 & \large 0.54 & \large 0.33 \\
       & & \large Avg & \large 0.76 & \large 0.68 & \large 0.56 & \large 0.76 & \large 0.76 & \multicolumn{1}{c|}{\large 0.56} & \large 0.78 & \large 0.74 & \large 0.59 & \large \textbf{0.69} & \large 0.60 & \multicolumn{1}{c|}{\large 0.50} & \large 0.65 & \large \textbf{0.59} & \large 0.43 \\
        \cmidrule{2-18}
       & \multirow{3}{*}{\begin{tabular}[c]{@{}c@{}} \textbf{PEMMA (DoRA)}\\ \textcolor{red}{(\textbf{\large params=0.005} $\Psi$)} \end{tabular}} & \large Tumor & \large0.78 & \large0.71 & \large0.71 & \large 0.81 & \large 0.82 & \multicolumn{1}{c|}{\large 0.78} & \large0.83 & \large0.82 & \large0.79 & \large 0.73 & \large 0.63 & \multicolumn{1}{c|}{\large 0.58} & \large 0.81 & \large 0.61 & \large0.68 \\
       & & \large Lymph & \large0.77 & \large0.71 & \large 0.65 & \large 0.75 & \large 0.71 & \multicolumn{1}{c|}{\large 0.40} & \large 0.76 & \large 0.71 & \large 0.54 & \large 0.61 & \large 0.59 & \multicolumn{1}{c|}{\large 0.43} & \large0.64 & \large0.56 & \large0.45 \\
       & & \large Avg & \large\textbf{0.78} & \large\textbf{0.71} & \large \textbf{0.68} & \large \textbf{0.78} & \large \textbf{0.77} & \multicolumn{1}{c|}{\large \textbf{0.59}} & \large \textbf{0.79} & \large  \textbf{0.76} & \large \textbf{0.66} & \large 0.67 & \large \textbf{0.61} & \multicolumn{1}{c|}{\large \textbf{0.51}} & \large\textbf{0.72} & \large 0.58 & \large\textbf{0.56} \\
       \bottomrule
     \end{tabular}
   }
 }
 \label{tab_results}
\end{table*}

\begin{figure*}[!htbp]
    \centering
    \includegraphics[width=0.9\linewidth]{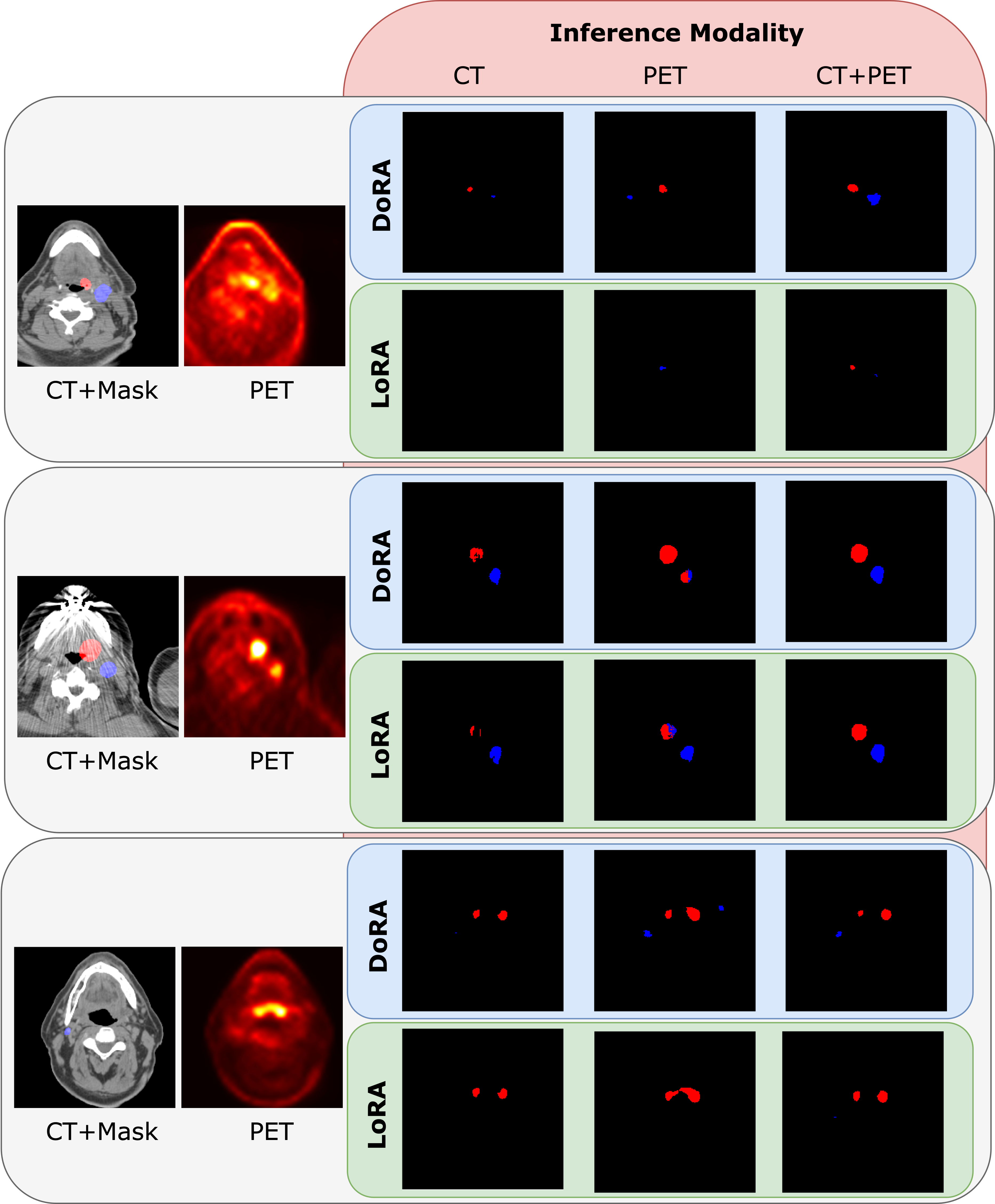}
    \caption{\textit{\textbf{Qualitative results of multi-modal adaptation stage:}} We review the detection/segmentation results of both gross tumor and lymph nodes. Especially, PEMMA generalizes well in organ segmentation and does not generate many false positives of tumors. However, it can be observed that overall DoRA outperforms LoRA-based PEFT.
    }
    \label{qual_adapt}
\end{figure*}

\begin{table}[ht]
\centering
\caption{We explore different dimensionality reduction techniques for UNETR while passing \(2N \times D\) in the encoder. We are faced with the need to keep the dimensions of the input to the decoder as \(N \times d\). Consequently, we compare three approaches: (1) alternating tokens from both modalities, (2) using only CT tokens, and (3) passing only CT tokens to the decoder. The latter approach yields superior Dice scores, motivating its adoption for subsequent experiments.}
\label{tab:dimensionality_reduction}
\begin{tabular}{lccc}
\toprule
\textbf{Metric}       & \textbf{Mix} & \textbf{CT only} & \textbf{PET only} \\
\midrule
Tumor Dice            & 0.47         & \textbf{0.63}    & 0.30              \\
Lymph Dice            & 0.20         & \textbf{0.57}    & 0.06              \\
Average Dice          & 0.34         & \textbf{0.60}    & 0.18              \\
\bottomrule
\end{tabular}
\label{UNETR_ablation}
\end{table}

\subsection{Experiment I: Segmentation}
We initially pre-trained the UNETR and Swin UNETR models using only the patients' CT scans data from centers CHUS, CHUV, CHUM, and CHUP, as shown in Table \ref{data_dist}. 
\subsubsection{Multi-modal Adaptation}
To evaluate the performance of the proposed method in a real-world clinical setting and assess the potential benefits of incorporating metabolic information from PET scans, the unimodal segmentation model was adapted to multi-modal model using registered CT and PET (CP) scans from the MDA center. A primary objective of these experiments was to determine the trade-off between adaptation performance and computational cost, aiming for maximum efficiency.

We evaluated the efficacy of our proposed method, PEMMA, compared to conventional fusion approaches, namely Early Fusion and Late Fusion, where we employed both Low-Rank Adaptation (LoRA) and Decomposed Orthogonal Rank Adaptation (DoRA) as part of our efficient adaptation. The architectural implementations of these fusion and adaptation techniques are detailed in the Methodology section.

To evaluate the multi-modal adaptation strategies during inference, we assessed the performance using CP data and single-modality inputs: only CT, and only PET, as shown in Table \ref{tab_results}. When both CT and PET modalities were available during inference, the Swin UNETR backbone consistently outperformed the UNETR backbone. Within the Swin UNETR framework, the PEMMA (DoRA) method achieved the highest average Dice score of 0.78, surpassing PEMMA (LoRA) at 0.76, with both approaches requiring only an additional $0.004\Psi$ parameters, highlighting their exceptional parameter efficiency. On the other hand, within the UNETR backbone, the Late Fusion method recorded the best average Dice score of 0.67, marginally outperforming Early Fusion and PEMMA (LoRA), both at 0.64; however, PEMMA (LoRA) matched this performance with far fewer trainable parameters ($0.08\Phi$) compared to Early Fusion ($1.0043\Phi$) and Late Fusion ($2\Phi$). In robustness evaluations, where models trained on combined CT and PET data were tested with only one modality, Swin UNETR’s PEMMA (DoRA) excelled, particularly when tested on CT alone (average Dice score of 0.71), and achieved the highest score with PET alone (0.68), despite a notable decline in performance compared to dual-modality evaluation. PEMMA (LoRA) showed similar trends, scoring 0.68 for CT and 0.56 for PET. For UNETR, Late Fusion struggled with single modalities (0.44 for CT and 0.47 for PET), while Early Fusion and PEMMA (LoRA) performed better, especially with PET (0.66 and 0.64, respectively), suggesting improved PET information retention.

\subsubsection{Continual Learning}
After adapting the model for multi-modal data (CT and PET) we proceeded to evaluate its continual learning capabilities by fine-tuning and testing it on entirely unseen data from new centers, namely, HGJ and HMR. Two distinct fine-tuning setups were investigated for each new dataset, as detailed below and summarized in Table~\ref{tab_results}:

\begin{itemize}
    \item \textbf{Fine-tuning with CT-only}: In this scenario, fine-tuning was conducted using only CT data from the new center (either HGJ or HMR), simulating a situation where PET data is unavailable. The models, previously adapted to the MDA dataset, were further fine-tuned exclusively on the CT scans from the respective new dataset.
    \item \textbf{Fine-tuning with CT+PET}: In this scenario, fine-tuning utilized both CT and PET data from the new center, reflecting a more comprehensive data availability. The models were fine-tuned using the combined CP data from the respective new dataset.
\end{itemize}

Following each fine-tuning process, the model's performance was assessed on the test split of the respective new dataset (HGJ or HMR) using combined (CP), CT-only, and PET-only inputs. This evaluation aimed to determine the effectiveness of fine-tuning under varying data availability conditions and to explore the models' cross-modal inference capabilities on the new data distribution.

\begin{figure}[!ht]
    \centering
    \includegraphics[width=1.0\linewidth]{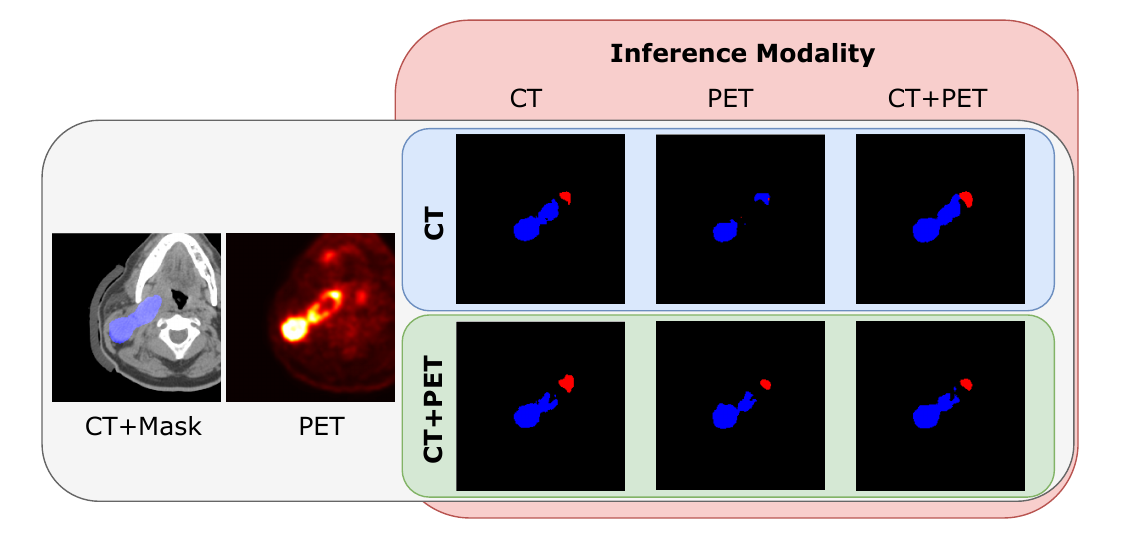}
    \caption{\textit{\textbf{Qualitative results of CL Task 1:}} We review the detection/segmentation results after finetuning using single modality i.e. CT and multi-modality i.e. CT+PET data from HGJ center. 
    }
    \label{fig_qual_HGJ}
\end{figure}

\subsubsection{Fine-tuning on HGJ}

When models were fine-tuned using only HGJ's CT data, the Swin UNETR model with PEMMA (DoRA and LoRA) exhibited notable adaptability. Specifically, after CT-only fine-tuning, PEMMA (DoRA) achieved high average Dice scores for CT-only inference (0.82), combined CP inference (0.81), and PET-only inference (0.78), indicating effective retention of multi-modal correlations learned during the initial adaptation. Similarly, PEMMA (LoRA) demonstrated comparable performance with scores of 0.81 for CP, 0.81 for CT, and 0.65 for PET. For the UNETR model, PEMMA (LoRA) adapted most effectively under CT-only fine-tuning, achieving scores of 0.82 for CP, 0.70 for CT, and 0.29 for PET, significantly outperforming the Early Fusion approach (0.76 for CP, 0.71 for CT, and 0.17 for PET) and the Late Fusion method (0.67 for CP, 0.45 for C, and 0.32 for P). These results underscore the efficacy of PEMMA in facilitating efficient adaptation to new data sources using only CT data, while preserving a degree of multi-modal inference capability acquired during the initial training on the MDA dataset.

Fine-tuning with both CT and PET data from HGJ generally yielded the best overall performance, particularly for CP and PET inference. The Swin UNETR model with PEMMA (DoRA) achieved the highest average Dice scores across all inference modalities: 0.79 for CP, 0.76 for CT, and 0.66 for PET. PEMMA (LoRA) was competitive, with scores of 0.78 for CP, 0.74 for CT, and 0.59 for PET. For the UNETR model, PEMMA (LoRA) showed strong results after CP fine-tuning, achieving 0.72 for CP, 0.71 for C, and 0.30 for P, which was comparable to or slightly better than the Early Fusion method (0.70 for CP, 0.68 for C, and 0.24 for P). Comparing the outcomes of CP versus CT-only fine-tuning on HGJ, utilizing both modalities during fine-tuning generally enhanced PET-only inference performance significantly for Swin UNETR PEMMA methods. Notably, CT-only fine-tuning proved surprisingly effective, particularly for CP and C inference, possibly attributable to the robust CT pre-training foundation.

\begin{figure}
    \centering
    \includegraphics[width=1.0\linewidth]{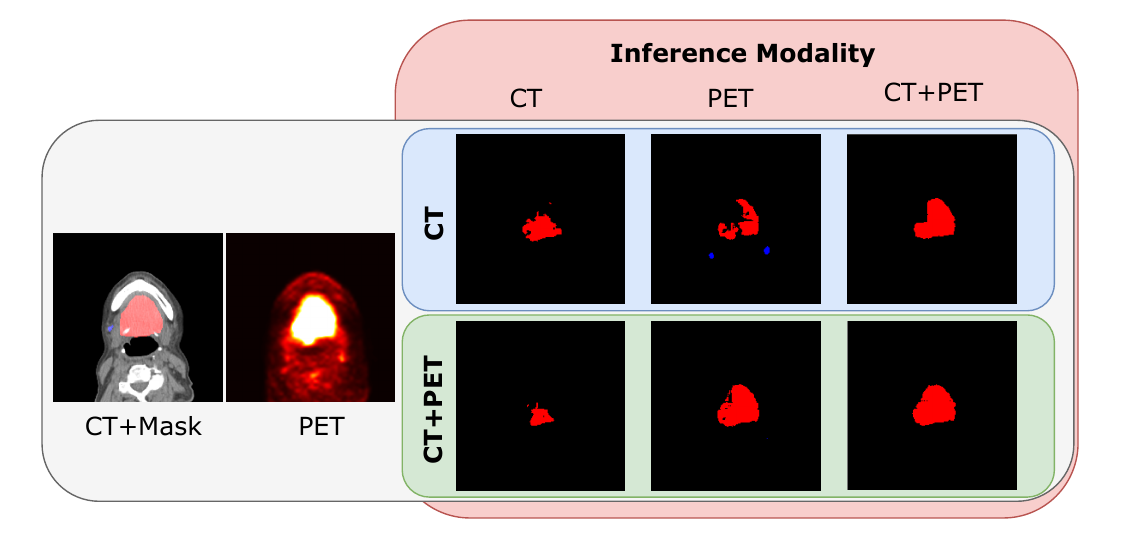}
    \caption{\textit{\textbf{Qualitative results of CL Task 2:}} We review the detection/segmentation results after finetuning using single modality i.e. CT and multi-modality i.e. CT+PET data from HMR center.}
    \label{fig_qual_HMR}
\end{figure}

\subsubsection{Fine-tuning on HMR}

Adapting to the HMR dataset using only CT data proved more challenging, potentially indicating a larger domain gap. Nevertheless, the Swin UNETR model with PEMMA (DoRA) maintained its lead, achieving average Dice scores of 0.73 for CP, 0.63 for CT, and 0.58 for PET. PEMMA (LoRA) followed closely with scores of 0.74 for CP, 0.66 for CT, and 0.66 for PET. For the UNETR model, performance declined considerably after CT fine-tuning on HMR compared to HGJ, with PEMMA (LoRA) achieving scores of 0.86 for CP, 0.38 for CT, and 0.34 for PET. The Late Fusion method struggled significantly, with scores of 0.68 for CP, 0.43 for CT, and 0.05 for PET.

Fine-tuning with HMR's CP data led to improved performance compared to CT-only fine-tuning, particularly for CP and PET inference. The Swin UNETR model with PEMMA (DoRA) achieved the best results: 0.72 for CP, 0.58 for CT, and 0.56 for PET, closely followed by PEMMA (LoRA) with 0.65 for CP, 0.59 for CT, and 0.43 for PET. For the UNETR model, PEMMA (LoRA) performed best among its group, achieving 0.63 for CP, 0.42 for CT, and 0.33 for PET, which was significantly better than the Early Fusion (0.60 for CP, 0.35 for CT, and 0.34 for PET) and Late Fusion (0.39 for CP, 0.40 for CT, and 0.18 for PET) methods. The comparison between CT-only and CP fine-tuning on HMR confirms that access to both modalities during the secondary adaptation phase is generally beneficial for maximizing performance, especially when facing a substantial domain shift.

\begin{figure*}
    \centering
    \includegraphics[width=1.0\linewidth]{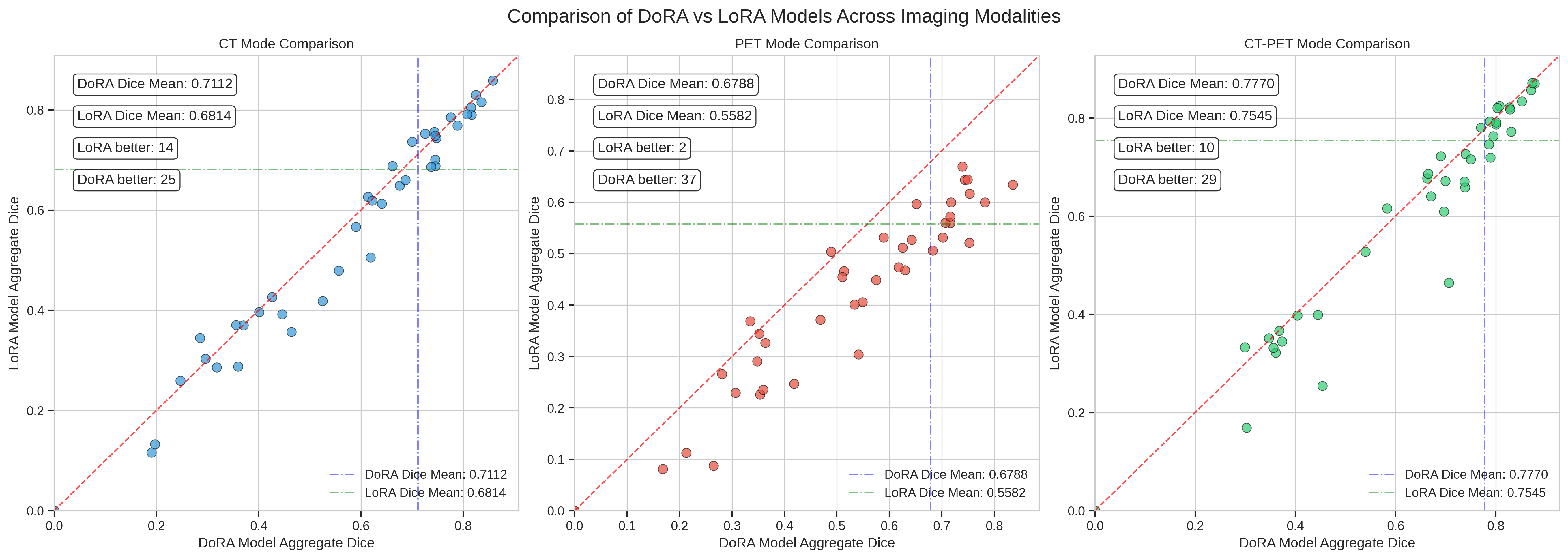}
    \caption{\textit{\textbf{Comparison of Swin UNETR performance under LoRA and DoRA adaptation methods: }} DoRA based adaptation outperforms LoRA based adaptation under both single modality adaptation (i.e. PET) and multi-modality adaptation (CT-PET).
    }
    \label{peft_comp}
\end{figure*}

\subsection{Experiment II: Prognosis}
To assess the flexibility of PEMMA in adapting to different tasks, we extend its application to multi-modal prognosis adaptation. We infer the prognosis experiments in three distinct settings: CT, CP, and CPT (CT+PET with textual EHR). These settings represent a progressive increase in the amount of available patient data, starting with imaging modalities and extending to the integration of textual EHR data, which provides richer context for patient outcomes. The addition of textual data in the CPT setting allows the model to harness detailed patient information beyond what is captured in imaging alone, thus offering a more comprehensive prognosis.

For preprocessing the EHR, each patient's
data is aggregated into a single sentence that is then processed by a pre-trained ClipMD text encoder \cite{clipmd} to generate high-dimensional text embeddings. Any missing information is omitted from the text, ensuring that the model only processes available data, thus avoiding the need for imputation strategies. An example text prompt for a patient is as follows: "\textit{This is a female head-and-neck cancer patient, 62 years old, weighing 84 kg, tobacco user, alcohol user, HPV positive, who underwent surgery, and received chemotherapy.}" 
The resulting text embedding is concatenated with the output of the Swin UNETR encoder and passed to the prognosis head to perform survival predictions. In addition to the PEFT layers in Swin UNETR for the imaging modalities, we employed LoRA and DoRA in the query, key, value, and fully connected layers of the text encoder using a rank of 32 and alpha scaling of 128. We used the same center splits for prognosis multi-modal adaptation as in the segmentation experiments. We report the time-dependent concordant indices (C-index) of Antolini et al. \cite{antolini}, a metric commonly used in survival analysis in the presence of censoring (i.e., when the times-to-event, in this case, recurrence-free relapse, of some patients are known while for others they are not).

Table \ref{tbl_prognosis} demonstrates that as additional modalities become available, PEMMA effectively adapts to leverage the increased patient information. Importantly, this adaptability is not limited to imaging data; PEMMA also excels at incorporating textual information from EHRs with ease. As a result, the C-index improves progressively from 0.61 (CT) to 0.67 (CP) to 0.75 (CPT) using LoRA, and from 0.66 (CT) to 0.68 (CP) to 0.80 (CPT) using DoRA. When comparing the two parameter-efficient fine-tuning methods, DoRA consistently outperforms LoRA across all configurations, mirroring the trends observed in the segmentation experiments. This reinforces the advantages of DoRA as a more effective fine-tuning approach. PEMMA's ability to integrate both imaging and non-imaging data with minimal parameters highlights its potential for scalable, efficient model adaptation, making it a powerful tool for various medical tasks.


\begin{table}
\caption{Concordant index results for prognosis using PEMMA. CP=CT+PET; CPT=CT+PET+text (i.e., EHR).}\label{tbl_prognosis}
\begin{tabular*}{\tblwidth}{@{}LLLL@{}}
\toprule
  & \textbf{CT} & \textbf{CP} & \textbf{CPT} \\ 
\midrule
 PEMMA (LoRA) & 0.6129 & 0.6720 & 0.7473 \\
 PEMMA (DoRA) & 0.6559 & 0.6828 & 0.8011 \\
\bottomrule
\end{tabular*}
\end{table}

\section{Discussion}

Our proposed method, PEMMA, was evaluated on individual modalities (CT alone, PET alone) and the combined modality (CT+PET), comparing it to conventional Early and Late fusion techniques. Our results demonstrate that PEMMA performs equally well to conventional methods; however, PEMMA is at least \textbf{12x} more efficient. Furthermore, PEMMA’s ability to adapt to new data was evaluated through two training setups that simulate real-world medical scenarios. Particularly, data from HGJ and HMR centers were used such that the pre-trained model was fine-tuned on CT data only (simulating scenarios where PET data are lacking), and the pre-trained model was fine-tuned using both CT and PET. Across experiments pertaining to these two scenarios, PEMMA outperformed both early and late fusion methods across various adaptation and inference approaches. This is highlighted in the notable improvements of minimum Average Dice scores of approximately 19\% and 28\% increases for two new datasets, respectively. Furthermore, the results shown underscore the strengths of Swin UNETR over UNETR within PEMMA. This observation can be attributed primarily to Swin UNETR’s advanced architecture, which incorporates the Swin Transformer. The use of a hierarchical Swin Transformer encoder allows Swin UNETR to extract features at multiple resolution levels, enhancing its ability to segment complex structures accurately.

Both PEFT techniques (LoRA and DoRA) enhance the model by gradually incorporating additional modality details while significantly reducing retraining requirements by 92\% due to their fewer trainable parameters. However, DoRA outperformed LoRA both quantitatively and qualitatively, as shown in Table \ref{tab_results} and Figure \ref{qual_adapt}, respectively. DoRA separates the weight matrix into magnitude and direction components, which allows it to mimic the learning behavior of full fine-tuning more closely than LoRA. This decomposition enhances both learning capacity and training stability \cite{dora}. Figure \ref{qual_adapt} shows an intricate example where both LoRA and DoRA have false positive segmentation; however, in most of the examples, DoRA has better performance. We show a detailed comparison of DoRA vs. LoRA in Figure \ref{peft_comp} on the test set of multi-modal adaptation stage. It can be observed that DoRA is consistently outperforming LoRA both in single and multi-modality scenarios. 

Moreover, our proposed approach proved to be advantageous for continual learning scenarios, where new tasks with varying data distributions (centers and modalities) are introduced over time. Due to the PEFT method's efficiency, only a small set of parameters requires updates per continual learning task, facilitating strong forward transfer and preventing negative backward transfer. As with the multi-modal adaptation results, we observed that DoRa consistently outperforms LoRA in the continual learning tasks as shown in Table \ref{tab_results}. The qualitative results in Figures \ref{fig_qual_HGJ} and \ref{fig_qual_HMR} illustrate how training modality (CT-only vs CT+PET) impacts segmentation performance when tested on different inference modalities (CT, PET and CT+PET). In Figure \ref{fig_qual_HGJ}  (CL Task 1, HGJ center), models trained on CT alone (top row) struggle to accurately segment structures when tested on PET data, with incomplete or misaligned masks (e.g., smaller red/blue regions). Conversely, models trained on CT+PET (bottom row) produce more precise and comprehensive segmentations across all modalities, indicating better adaptability. Similarly, Figure \ref{fig_qual_HMR} (CL Task 2, HMR center) shows that multi-modality training enhances robustness. While single-modality-trained models may miss critical regions (e.g., fragmented masks in PET inference), CT+PET-trained models consistently capture target structures regardless of the input modality. These findings highlight the importance of leveraging combined CT+PET data during adaptation to ensure reliable performance when deploying the model in diverse clinical settings with varying imaging protocols. 

Beyond segmentation, we also evaluate the ability of PEMMA to be extended to another challenging medical task. Using only the encoder of Swin UNETR, PEMMA can be efficiently adapted to perform prognosis using CT scans only and CT+PET together. Moreover, when extended to include electronic health records (EHR), the model further improves its adaptability by incorporating rich, non-imaging data. This integration allows PEMMA to leverage both imaging and textual patient information, enhancing its performance as more data modalities are added. These findings highlight PEMMA's flexibility in handling multi-modal data and its ability to provide robust prognosis predictions even when the available data varies across patients, offering significant advantages in terms of flexibility and computational efficiency for real-world clinical applications.

\section{Conclusion}
In this work, we introduced PEMMA, a novel and efficient multi-modal model adaptation framework for medical image segmentation and prognosis prediction, demonstrating its dual efficacy in both tasks through the integration of imaging and language modalities. By leveraging minimal parameter training and small-scale datasets, PEMMA achieves robust performance while addressing practical constraints in clinical settings, such as limited data availability and computational resources. Furthermore, the framework exhibits strong generalization capabilities via continual learning, mitigating catastrophic forgetting during sequential adaptation to new domains or centers, thereby enabling reliable deployment across diverse clinical environments. Unlike static multi-modal architectures that rely on resource-intensive decoders or full model retraining, PEMMA dynamically integrates heterogeneous modalities and tasks, offering a scalable solution for evolving clinical needs. However, the current requirement for pre-registered medical images remains a limitation, as unregistered data—a common challenge in real-world scenarios—poses unresolved complexities. Future work will focus on extending PEMMA to handle unregistered multi-modal inputs, which would significantly enhance its adaptability in settings where image registration is impractical or unavailable. Overall, PEMMA represents a critical advancement in multi-modal medical AI, bridging the gap between resource-efficient adaptation and clinical applicability.




\bibliographystyle{cas-model2-names}

\bibliography{cas-refs}





\end{document}